\pdfoutput=1

\documentclass[11pt]{article}

\usepackage[]{EACL2023}

\usepackage{times}

\usepackage{multirow}
\usepackage{graphicx}

\usepackage{subfigure}

\usepackage[T1]{fontenc}

\usepackage[utf8]{inputenc}

\usepackage{microtype}

\usepackage{inconsolata}

\title{VL-CheckList: Evaluating Pre-trained Vision-Language Models with Objects, Attributes and Relations}

 \author{Tiancheng Zhao$^{1,2}$, Tianqi Zhang$^{3}$, Mingwei Zhu$^{3}$, Haozhan Shen$^{3}$, \\
  \textbf{Kyusong Lee$^{1,2}$, Xiaopeng Lu$^{1,2}$, Jianwei Yin$^{3}$} \\
  Om Research Lab, Binjiang Institute of Zhejiang University$^{1}$ \\
  Linker Technology Research Co. Ltd$^{2}$ \\
  College of Computer Science and Technology, Zhejiang University$^{3}$ \\
  \texttt{\{tianchez, kyusongl\}@zju-bj.com}, \texttt{lu\_xiaopeng@hzlh.com}\\
  \texttt{\{zhang\_tq, zhumw, cnfighting, zjuyjw\}@zju.edu.cn}
  }

\begin{document}
\maketitle
\begin{abstract}
Vision-Language Pretraining (VLP) models have recently successfully facilitated many cross-modal downstream tasks. Most existing works evaluated their systems by comparing the fine-tuned downstream task performance. However, only average downstream task accuracy provides little information about the pros and cons of each VLP method, let alone provides insights on how the community can improve the systems in the future. Inspired by the CheckList~\cite{ribeiro2020beyond} for testing natural language processing, we exploit VL-CheckList, a novel framework to understand the capabilities of VLP models. The proposed method divides the image-texting ability of a VLP model into three categories: objects, attributes, and relations, and uses a novel taxonomy to further break down these three aspects. We conduct comprehensive studies to analyze seven recently popular VLP models via the proposed framework. Results confirm the effectiveness of the proposed method by revealing fine-grained differences among the compared models that were not visible from downstream task-only evaluation. Further results show promising research direction in building better VLP models. Our data and code are available at ~\url{https://github.com/om-ai-lab/VL-CheckList}\;. 
\end{abstract}

\section{Introduction}

\begin{figure*}[h!]
\centering
  \includegraphics[width=14cm]{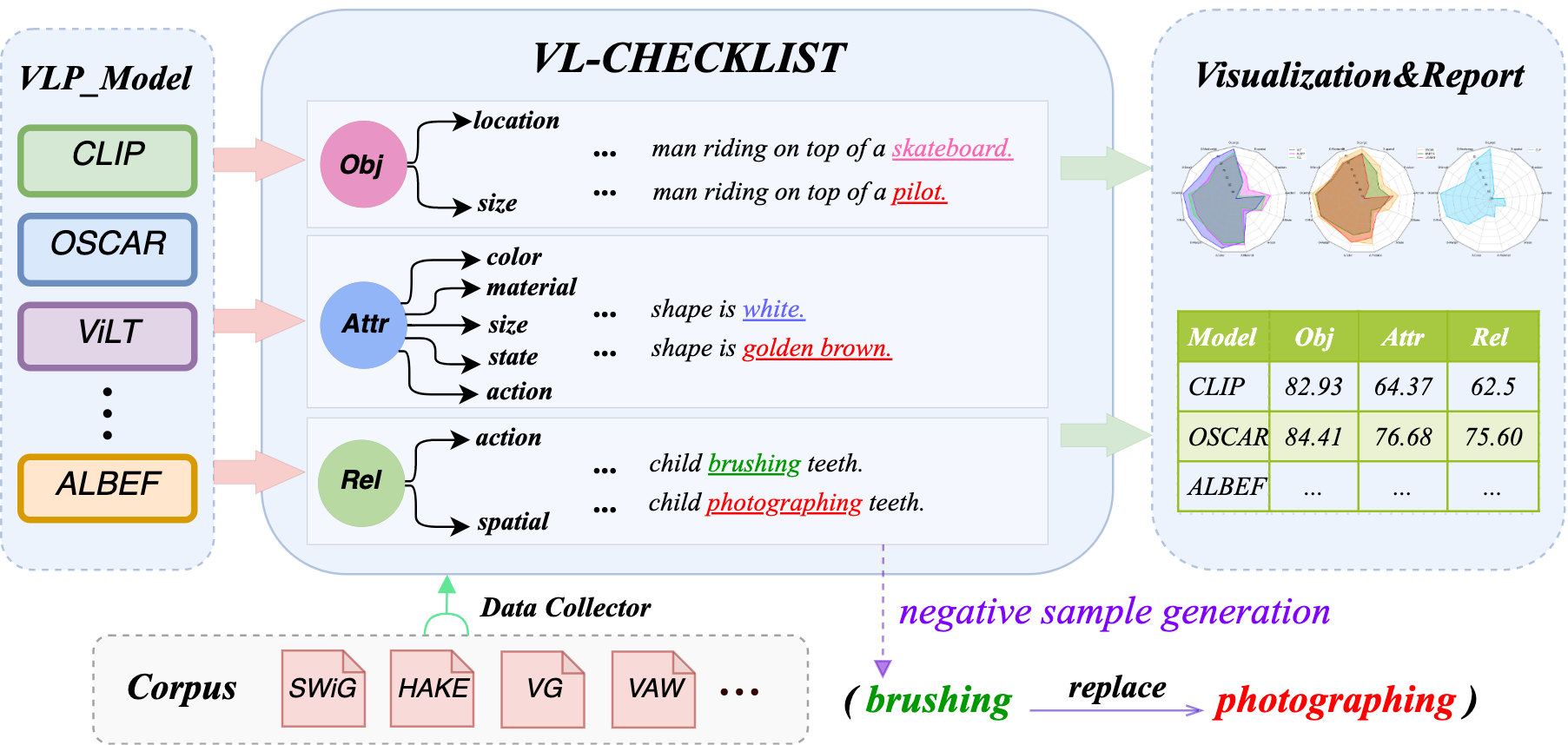}
  \caption{Overview of the proposed VL-CheckList}
  \label{fig:outline}
\end{figure*}

Vision-Language Pretraining (VLP) is a fundamental task for multimodal machine learning research. Recently, VLP has rapidly progressed~\cite{li2020oscar,radford2021clip,li2021align,kim2021vilt}, thanks to the emergence of multimodal transformers~\cite{vaswani2017attention} and the availability of large paired image-text corpora~\cite{sharma2018conceptual,changpinyo2021conceptual}. Many proposed VLP models have helped to achieve the state-of-the-art performance of a variety of downstream multimodal tasks, ranging from visual QA~\cite{lu2019vilbert}, multimodal retrieval~\cite{lu2021visualsparta} to visual grounding~\cite{kamath2021mdetr} and many others. On the other hand, the current defacto method to evaluate a VLP model is by comparing its fine-tuned downstream tasks performance. However, benchmark VLP models solely based on a group of downstream tasks have a number of limitations: 

\textbf{(1) Poor Interpretability}: a downstream task is complex and depends on many intertwined capabilities; thus, it only provides a black box score and is difficult to decipher. For example, it is unclear how to improve a VLP model that excels at Visual QA, but performs poorly in image retrieval. \textbf{(2) Incomparable Results}: different works may choose different tasks for evaluation, making it hard to compare. This is due to some VLP models are not compatible with certain tasks, such as CLIP~\cite{radford2021clip} cannot be fine-tuned for visual QA directly. \textbf{(3) Bias in Data}: downstream data distribution is not comprehensive, such that real-world performance may be overestimated. Also, it is unknown if a model is robust against input noise, e.g., replacing a verb by its synonymy.

To address the above limitations, this paper introduces VL-CheckList, an explainable framework that comprehensively evaluates VLP models, facilitates deeper understanding and inspires new ideas for improvement. The core principles of VL-CheckList are: (1) evaluate a VLP model's fundamental capabilities instead of performance on applications (2) disentangle capabilities into relatively independent variables that are easier to analyze. Specifically, we choose Image-Text-Matching (ITM) as the primary evaluation target since it is perhaps the most effective pretraining objective that appears in all VLP methods~\cite{Li2019VisualBERTAS,li2020oscar,radford2021clip,li2021align}. We then propose a taxonomy that divides the capabilities of VLP systems into three categories: object, attribute and relation. Each aspect is further divided into more fine-grained variables, e.g. attribute is composed of color, material, size, etc. Then, a linguistic-aware negative sampling strategy is proposed to create ''hard negative'' that challenges a VLP model's discriminative power against minor changes in the input space. Lastly, VL-CheckList is implemented as a toolbox that is easy for everyone to plug into their evaluation pipeline. 

The proposed method is validated by analyzing 7 popular VLP models, including dual-encoder models CLIP~\cite{radford2021clip}, region-based VLP models, e.g. OSCAR~\cite{li2020oscar} and end-to-end VLP systems, e.g. ViLT~\cite{kim2021vilt}. Four datasets: VG~\cite{Krishna2016VisualGC}, SWIG~\cite{Pratt2020Swig}, VAW~\cite{Pham2021vaw}, and HAKE~\cite{Li2019HAKEHA} are adopted to generate capability-specific evaluation test set. Experiment results reveal several interesting insights about these models that are difficult to obtain from down-stream task scores. 
In short, the contributions of this work are four folds:
\begin{itemize}
    \setlength\itemsep{0.01em}
    \item We present VL-CheckList, a novel explainable framework that generates fine-grained and disentangled evaluation reports about VLP models. 
    \item The proposed method profiles VLP models by testing the ability to understand an object, attribute, and relations between a text prompt and a given image. 
    \item Experiment results that show the superiority of the proposed framework by providing several promising insights on seven popular VLP systems. 
    \item Open-source package of VL-CheckList that is easy to be used by the community.
\end{itemize}

\section{Related Work}
Two lines of research have closely related to this work:

\noindent\textbf{Machine Learning Evaluation Tools: }
Machine learning (ML) models are often tested on benchmark datasets~\cite{rajpurkar2016squad,bowman2015large,wang2018glue}. However, without a comprehensive analysis, it is difficult to understand the strengths and weaknesses of a model. Recent studies show even the state-of-the-art systems may still be insufficient in real-world applications~\cite{ribeiro2020beyond}. Thus, researchers have attempted to evaluate ML models with more fine-grained details.

One of the popular tools for the qualitative analysis of natural language processing (NLP) is CheckList~\cite{ribeiro2020beyond}. It evaluates general linguistic capabilities and shows weaknesses in several state-of-the-art NLP models. Dynabench~\cite{kiela2021dynabench} was proposed to generate dynamic benchmark datasets. It overcomes the problem that the existing benchmark fails to cover simple fundamental linguistic challenges. In computer vision, the Vision CheckList was proposed to help system designers to understand model capabilities~\cite{du2022vision}. They offered a set of transformation operations to generate diverse test samples of different test types, such as rotation, replacing image patches, blur, and shuffle. TIDE~\cite{bolya2020tide} is a tool to analyze the errors of object detection. It defines critical error types and shows a comprehensive analysis. 

\noindent\textbf{Analysis of VLP Models: }
Prior work has also attempted to gain a deeper understanding of VLP models. \citet{hendricks2021decoupling} showed the influence of multimodal representations based on three variables: datasets, the attention mechanism, and objective function. They found that 1) the qualify of the image captions is crucial for multi-modal training 2) having multi-modal attention with a smaller model is better than deep models without multi-modal attention 3) the image loss (masked region modeling) is not very critical for the performance. \citet{hendricks2021probing} investigated the capability of the multi-modal transformers to understand verbs. They observed that understanding verb is harder than objects and subjects. \citet{dou2022empirical} conducted the empirical analysis on each component of a VLP model: vision encoder, text encoder, fusion, objective, and model architecture. They reported the experiment results in different parts to provide a helpful recipe for obtaining better VLP training.

This paper differs from the prior art by being the first work (to the best of our knowledge) that analyzes VLP models from three angles: object, attribute, and relations, and provides a large-scale study that tests seven latest models on four corpora.

\section{VL-CheckList}
An intuitive approach to evaluate multi-modal systems is to check if a model correctly predicts alignment between different modalities. We choose image-text matching (ITM) to check the alignment between vision and language for the following reasons. Specifically, ITM is defined as the function that outputs the probability of an image $i$ is matched to a sentence $t$.  First, the ITM loss is often universal in all VLP models~\cite{li2020oscar}, so we can compare directly without any modification of the models. Second, the ITM is also model agnostic and applies to all fusion architectures. Thus, we exploit the ITM to fairly compare the VLP models without finetuning them on downstream tasks.

In short, the overall pipeline of VL-CheckList is described as the following (Figure~\ref{fig:outline})
\begin{enumerate}
    \setlength\itemsep{0.01em}
    \item We transform image-text paired datasets by categorizing samples into three aspects: Object, Attribute, and Relation. 
    \item The paired text for each image is rewritten to generate negative samples for each aspect.
    \item We use the ITM head of a VLP model to distinguish between the positive and the negative text is given an image. 
    \item Generate a comprehensive analysis report about the models for each aspect. 
\end{enumerate}

\subsection{Taxonomy of VL-CheckList}
\begin{figure}[h!]
\centering
  \includegraphics[width=7cm]{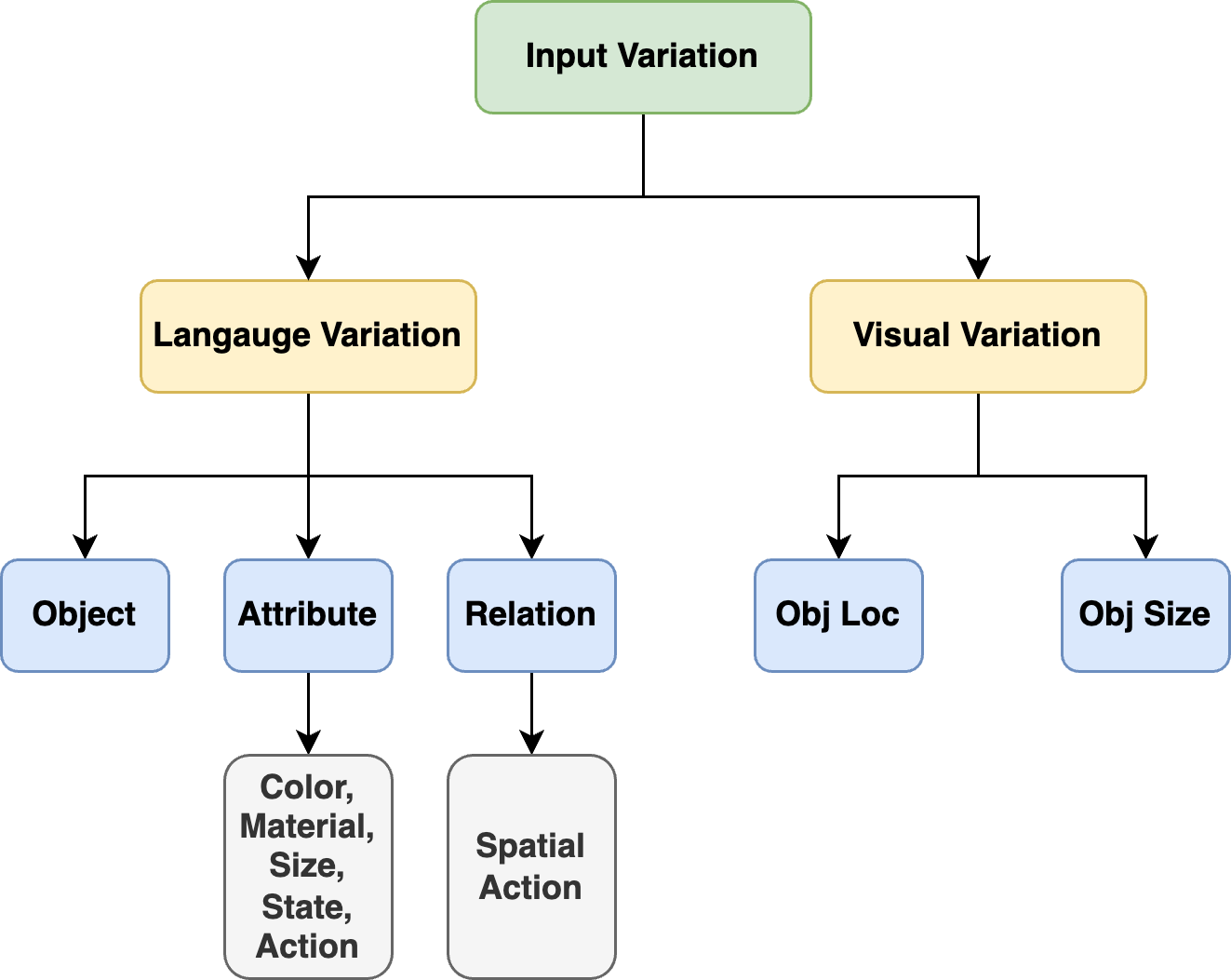}
  \caption{The proposed evaluation taxonomy}
  \label{fig:taxonomy}
\end{figure}
The evaluation types are often selected based on common mistakes or frequent usage. Based on the common issues in VLP models, the proposed framework places the three input properties (object, attribute, and relation) as the top layer of the evaluation taxonomy. 

\textbf{Object: }
A robust VLP model is supposed to recognize whether or not the objects mentioned in a text exist in the image. Therefore, if we replace objects in the correct text with some other random noun phrases, a VLP model should give it a lower ITM score than the original sentence. Furthermore, a robust VLP model should recognize objects' existence, regardless of their location and sizes. Thus, we further evaluate the robustness Object ITM by testing location invariance (e.g, center, middle, and margin) and size invariance (e.g., small, large, medium), specifically:

loc(x, y)=
$\left\{ 
  \begin{array}{ c l }
    center & \quad \textrm{if } \frac{y}{x} \leq \frac{1}{3} \\
    mid & \quad \textrm{if } \frac{1}{3} < \frac{y}{x} \leq \frac{2}{3} \\
    margin                 & \quad \textrm{otherwise}
  \end{array}
\right.$

where, $x$ is the half-length of the diagonal of the full image $x =\frac{\sqrt{w^2+h^2}}{2}$. and $y$ is the distance between its central point and the central point of the full image. 

To get the size of an object, we use the object area information (i.e., the bounding box of height multiplies the width).

size(x)=
$\left\{ 
  \begin{array}{ c l }
    small & \quad \textrm{if } area \leq S \\
    medium & \quad \textrm{if } S < area \leq M \\
    large                 & \quad \textrm{otherwise}
  \end{array}
\right.$

where, $area = w*h$, $S$ denotes small size and $M$ is the medium size. We set $S=1024, M=9216$ in this paper. 

\textbf{Attribute:}
Determining specific attributes for any object is very challenging. A strong VLP ITM head should assign a lower score if the correct attribute in the text is replaced. The attribute generally contains color, material, size, state, and action. 
\begin{itemize}
    \setlength\itemsep{0.01em}
    \item Size: replace the size expression like small, big, and medium with another  (e.g., There is a big apple vs. there is a small apple)
    \item Material: replace a material word in the sentence (e.g., a metal box vs. a wood box)
    \item State: replace the state expression, such as cleanliness and newness (e.g., a man with dry hair vs. a man with wet hair).
    \item Action: replace the action-related word in the text (e.g., a standing person vs. a sitting person).
    \item Color: replace the color word in the text (e.g., A red apple is on the table vs. A green apple is on the table)
\end{itemize}

\textbf{Relation: }
Relation cares about the interaction between two objects. It covers replacing the predicate in a triple (e.g., subject, predicate, object), where the subject and object are both objects in the image. A strong VLP ITM head should assign a higher score to text matching the pair-wise object interaction. Further, we divide prediction into spatial and action. If a predicate is one of the spatial prepositions (e.g., in, on, at, etc.), it is sub-categorized as 'spatial'; otherwise, it is labeled 'action'. 
\begin{itemize}
    \setlength\itemsep{0.01em}
    \item Spatial: If a model can predict spatial relation between two objects (e.g, \textless cat, on, table\textgreater vs. \textless cat, under, table\textgreater).
    \item Action: If a model can predict other relation than a spatial preposition, usually action verbs like run, jump, kick, eat, break, cry, or smile (e.g., \textless cat, catch, fish\textgreater vs. \textless cat, eat, fish\textgreater).
\end{itemize}

\begin{table}[]
\begin{tabular}{|l|l|l|}
\hline
\textbf{Name}  & \textbf{Size} & \textbf{Adopt to} \\ \hline
VG   & 108K  & Objects, Attribute, Relation   \\ \hline
VAW & 72K & Attribute  \\ \hline
HAKE & 104K & Objects, Relation \\ \hline
SWiG  & 126K & Objects, Relation  \\ \hline
\end{tabular}
\caption{A summary of the used evaluation datasets.}
\label{tbl:data-summary}
\end{table}

\begin{figure*}[h!]
\centering
  \includegraphics[width=0.7\textwidth]{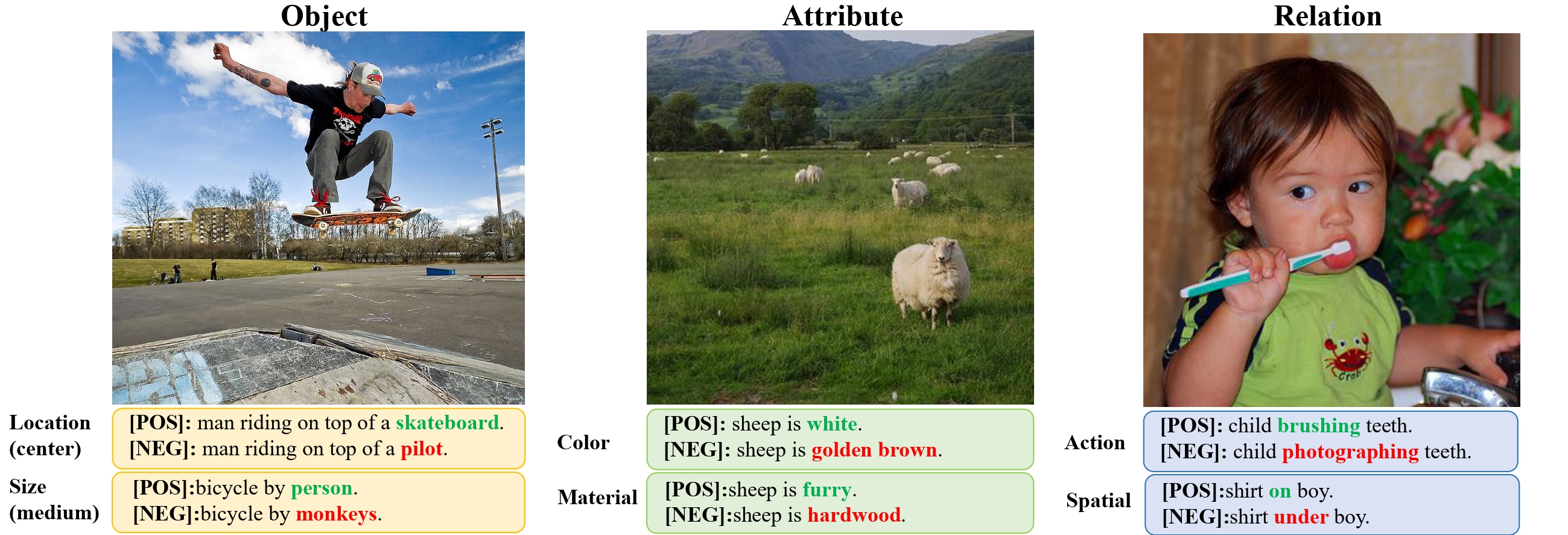}
  \caption{Example positive and negative samples for each evaluation aspect.}
  \label{fig:negative}
\end{figure*}

\subsection{Negative sampling generation}
The negative sampling generation is a set of transformations on the original text descriptions. The proposed VL-CheckList focuses on a directional expectation test, in which the label is expected to change in a certain way. For example, when there is a black bear in the photo and the text description is "A black bear is holding a stick". We can transform several variations (e.g., (a black bear $\rightarrow$ a red bear), (a stick $\rightarrow$ an apple), (holding $\rightarrow$ throwing), etc). The negative sampling strategy is the essential step for unbiased evaluations. 
To generate hard negative examples, we use the structured text description datasets such as Visual Genome (VG)~\cite{krishna2017visual}, SWiG~\cite{Pratt2020Swig}, and Human Activity Knowledge Engine (HAKE)~\cite{Li2019HAKEHA}. The VG provides attributes, relationships, and region graphs which can make a hard negative sample by replacing one attribute in the relation in the image. The SWiG dataset provides structured semantic summaries of images with their roles such as agent and tool. We generate hard negative samples by replacing one of the roles in the text description to mismatch with the image. HAKE dataset provides the relationship between instance activity and body part states (e.g., "head" inspect rear view, "right hand" hold wheel, "hip" sit on chair seat). 

For VG dataset, we first assign each attribute, object, and relation to the closet type by cosine similarity from sentence transformers\footnote{https://www.sbert.net/}. For objects and relationships, we randomly sample a corresponding instance with a cosine similarity threshold of 0.5. For attribute, we randomly sample a corresponding instance from the same attribute class with a cosine similarity threshold of 0.5. Some of the samplings are not very suitable. For some inappropriate captions pairs, we manually modified them.

\subsection{Visualization \& Metric}
\textbf{Visualization: } The final report is to show the models' capabilities. The previous NLP CheckList shows the visual summary table which includes a failure rate with examples of each evaluation type. The vision checklist also shows the interpretable components that contributed to the prediction and the degree to which each feature contributed to the prediction and attention map using Grad-CAM~\cite{selvaraju2017grad}. 
Our VL-CheckList is designed to comprehensively analyze a VLP model and output useful intermediate metrics for analysis and visualization. We provide the performance quantitative table and the radar chart based on the evaluation type taxonomy: object, attribute and relation and visualized insightful chart (Figure~\ref{fig:spider}) and the attention map (Figure~\ref{heat-map-car} and~\ref{fig:heat-map-working}).

\noindent\textbf{Metric: }We return the model output scores between the text description and the generated negative samples. If the model score on the original text description is higher than the score on the generated negative samples, we regard it as positive output. We object to accuracy (acc) as the following equation. 
\begin{equation}
acc = \frac{\sum_{i=0}^{i<n} f(x^p_i, x^n_i)}{N}
\end{equation}
where, $ f(x^p_i, x^n_i) = 1$ if $p(x^p_i|I_i) > p (x^n_i|I_i)$, otherwise 0. $x^p_i$ denotes a positive sample of $i^{th}$ data. $x^n_i$ means a positive sample of $i^{th}$ data. The N is the total number of pairs of positive and negative samples. $I_i$ is $i^{th}$ image data.

\section{Experiment Setup}
We compare 7 state-of-the-art VLP methods via the proposed method. They are:

\begin{figure*}[]
\centering
\includegraphics[scale=0.25]{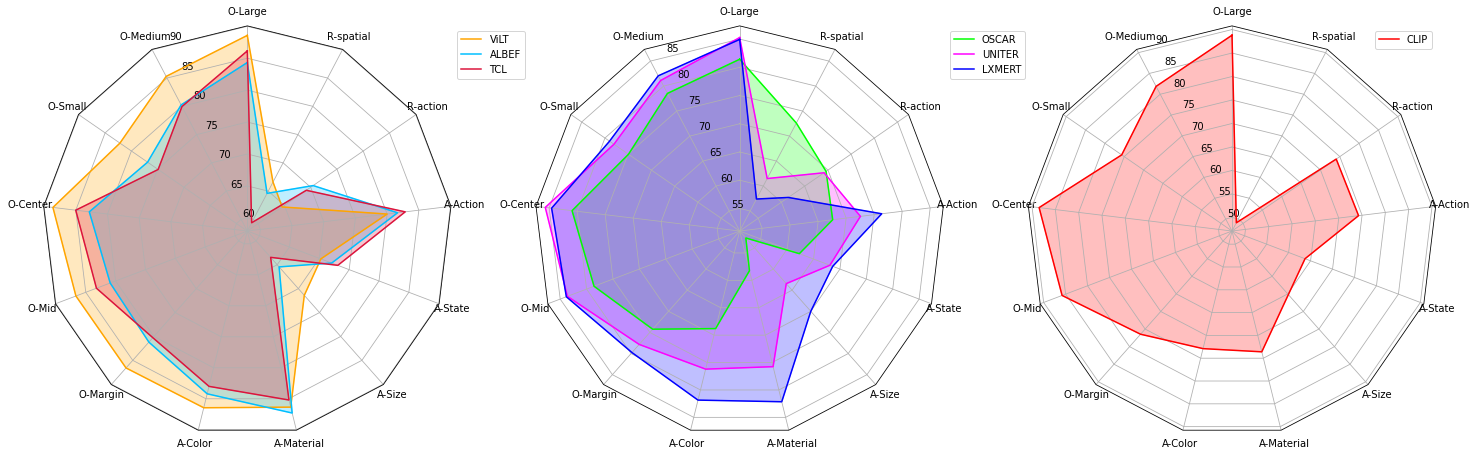}
\caption{Radar charts for the compared models.}
\label{fig:spider}
\end{figure*}

\subsection{Dual-Encoders}
\noindent\textbf{CLIP}~\cite{radford2021clip} is trained on a dataset of 400 million (image, text) pairs with contrastive objectives. CLIP has a simple model where two transformers encode the image and text independently. The ITM score is simple cosine distance between the two embeddings. Despite of CLIP's simple structure, it has achieved superior performance in many downstream tasks.

\subsection{Region-based Cross-Attention Models}
This type of models share the common trait that the input image is passed through a off-the-shelf object detectors to obtain salient objects and region features. They are:

\noindent\textbf{LXMERT}~\cite{tan2019lxmert} is a two-stream model with three transformers. An object relation encoder encodes the input objects and their location; a BERT encoder encodes the text; and a cross-modal encoder fuses the two modalities. The model is pretrained with ITM and three other objectives. 

\noindent\textbf{UNITER}~\cite{chen2020uniter} is a single-stream model that is pre-trained with four objectives. UNITER is different from LXMERT with its single-stream design where a shared transformer fuses the visual and language information in early stage.

\noindent\textbf{OSCAR}~\cite{li2020oscar} is a single-stream model trained with masked token loss and constrastive loss on 6.5 million text-image-tag triplet. Its unique features are: (1) have a better object detector~\cite{zhang2021vinvl} that recognizes more types (2) include object tags with image and text as input features. 

\subsection{End-to-end Cross-Attention Models}
End-to-end models do not depend on object detectors by encoding images directly. 

\noindent\textbf{ViLT}~\cite{kim2021vilt} is an efficient single-stream model. The input image is divided into patches and concatenated with the words. Then a transformer encodes the combined sequence. ViLT is the first VLP model that achieves competent performance without using region futures.

\noindent\textbf{ALBEF}~\cite{li2021align} is a two-stream model. The text and image are encoded via RoBERTa~\cite{liu2019roberta} and ViT~\cite{dosovitskiy2020image} respectively. Then a cross-modal transformer is used to fuse the two modalities. In addition, to learn better representations from noisy image-text pairs, ALBEF learn from pseudo-targets produced by the momentum model.

\noindent\textbf{TCL}~\cite{yang2022tcl} is an improved version of ALBEF. TCL introduces three contrastive modules: cross-modal alignment (CMA), intra-modal contrastive (IMC), and local MI maximization (LMI), which respectively aims to maximize the mutual information between the matching image and text, maximizes the global mutual information.

We used four corpora, VG~\cite{krishna2017visual}, SWiG~\cite{Pratt2020Swig}, VAW~\cite{Pham2021vaw} and HAKE~\cite{Li2019HAKEHA} to build benchmark dataset for each capability test in the proposed framework (Table~\ref{tbl:data-summary}).

\begin{table*}[]\centering
\begin{tabular}{c|c|cccc|c}
\hline
Model Type                    & Models & Average (Model) & Object         & Attribute & Relation & Average (Type)                  \\ \hline
\multirow{3}{*}{REGION} & LXMERT & 72.35           & 82.36          & 77.12     & 57.57    & \multirow{3}{*}{71.43}          \\
             & UNITER & 72.63          & 81.94          & 71.31          & 64.65          &       \\
             & OSCAR  & 69.31 & 78.10 & 61.28          & \textbf{68.54} &       \\ \hline
\multirow{3}{*}{E2E}          & ViLT   & \textbf{76.82}  & \textbf{86.32} & \textbf{80.36}     & 63.79    & \multirow{3}{*}{\textbf{75.67}} \\
             & ALBEF  & 75.62          & 81.08          & 79.33          & 66.45          &       \\
             & TCL    & 74.57          & 81.58          & 78.27 & 63.85          &       \\ \hline
DUAL & CLIP   & 71.65          & 82.83       & 67.93          & 64.19          & 71.65 \\ \hline
\end{tabular}
\caption{Overall performance (E2E means end-to-end models, which are ViLT, ALBEF and TCL in our paper, while Region means Region-based approach including OSCAR, UNITER and LXMERT, and finally, DUAL means Dual-Encoder, CLIP)}
\label{all}
\end{table*}
\begin{table*}\centering
\begin{tabular}{c|c|c|ccc|ccc|c}
\hline
\multirow{2}{*}{Type}   & \multirow{2}{*}{Model} & \multirow{2}{*}{Avg} & \multicolumn{3}{c|}{Size}      & \multicolumn{3}{c|}{Location} & \multirow{2}{*}{Avg}            \\ \cline{4-9}
     &        &       & Large          & Medium         & Small & Center & Mid   & Margin &       \\ \hline
\multirow{3}{*}{REGION} & LXMERT                 & 82.36                & 84.94 & 82.05 & 79.01          & 84.56    & 83.84    & 79.77   & \multirow{3}{*}{80.80}          \\
     & UNITER & 81.94 & 85.27          & 81.12          & 78.06 & 85.69  & 83.64 & 77.84  &       \\
     & OSCAR  & 78.10 & 81.43          & 78.53 & 74.93 & 80.91  & 78.59 & 74.24  &       \\ \hline
\multirow{3}{*}{E2E}    & ViLT                   & \textbf{86.32}       & 88.58 & \textbf{85.29} & \textbf{82.18} & \textbf{88.61}    & \textbf{86.65}    & \textbf{86.59}   & \multirow{3}{*}{\textbf{82.99}} \\
     & ALBEF  & 81.08 & 84.33          & 80.25          & 76.92 & 82.87  & 80.91 & 81.19  &       \\
     & TCL    & 81.58 & 86.15          & 79.96          & 74.91 & 85.01  & 83.21 & 80.26  &       \\ \hline
DUAL & CLIP   & 82.83 & \textbf{88.77} & 81.82          & 75.65 & 88.48  & 85.80 & 76.47  & 82.83 \\ \hline
\end{tabular}
\caption{Evaluation of Object}
\label{obj}
\end{table*}

\begin{table*}[]\centering
\setlength{\tabcolsep}{1mm}{
\begin{tabular}{c|c|c|cc|ccc|ccc|c}
\hline
\multirow{2}{*}{Type}   & \multirow{2}{*}{Model} & \multirow{2}{*}{Avg} & \multicolumn{2}{c|}{Text level} & \multicolumn{6}{c|}{Image level} & \multirow{2}{*}{Avg}\\
\cline{4-11}
     &        &                & Act            & Spa  &  Large & Medium & Small & Center & Mid & Margin     \\ \hline
\multirow{3}{*}{REGION} & LXMERT                 & 57.57                & 61.47     & 57.4  &  57.70 & 57.36 & 53.37& 57.37 &56.43 &52.04 & \multirow{3}{*}{63.59} \\
     & UNITER & 64.65          & 69.17          & 61.5 & 64.73 & 66.17 &60.31 & 63.33 & 63.70 & 65.60      \\
     & OSCAR  & \textbf{68.54} & 69.6          & \textbf{72.6} &   67.65 &65.65 &61.50 &70.70 & 67.00 & 63.43    \\ \hline
\multirow{3}{*}{E2E}    & ViLT                   & 63.79                & 64.57     & 66.6  & 62.37 & 62.97 & \textbf{62.36} & 60.80 & 62.33 & 61.20   & \multirow{3}{*}{\textbf{64.70}} \\
     & ALBEF  & 66.45          & 70.47          & 64.6 &  65.17 & \textbf{67.06} &61.59 & 65.57 & 65.73 & \textbf{67.13}     \\
     & TCL    & 63.85          & 69.17          & 59.4 &  66.90 & 62.34 & 58.17 & 66.57 & 63.97 & 62.60
     \\ \hline
DUAL & CLIP   & 64.19          & \textbf{74.07} & 49.1  & \textbf{71.07} & 65.17 & 60.74 & \textbf{71.33} & \textbf{68.40} & 64.03 & 64.19  \\ \hline
\end{tabular}
}
    \caption{Evaluation of Relation (Act denotes Action and Spa denotes Spatial)}
    \label{relation}
\end{table*}



\begin{table*}[]
\centering
\setlength{\tabcolsep}{1mm}{
    \begin{tabular}{c|c|c|ccccc|ccc|ccc|c}
\hline
\multirow{2}{*}{Type} &
  \multirow{2}{*}{Model} &
  \multirow{2}{*}{Avg} &
  \multicolumn{5}{c|}{Text level} &
  \multicolumn{6}{c|}{Image level} &
  \multirow{2}{*}{Avg} \\ \cline{4-14}
     &        &       & Color & Mat   & Size  & State & Act &  Large & Medium & Small & Center & Mid & Margin      \\ \hline
\multirow{3}{*}{REGION} &
  LXMERT &
  77.12 &
  81.85 &
  82.15 &
  70.00 &
  68.63 &
  76.35 & 82.25 & 77.10 & 75.36 & 77.50 & 79.25 & 79.25 &
  \multirow{3}{*}{69.90} \\
     & UNITER & 71.31 & 76.20 & 75.75 & 63.45 & 68.10 & 72.55 &  74.90 & 69.45 & 68.49 & 72.05 & 73.30 & 70.25     \\
     & OSCAR  & 61.28 & 68.80 & 58.25 & 52.65 & 62.31 & 67.60 & 64.80 & 59.15 & 58.18 & 65.75 & 59.15 & 56.80      \\ \hline
\multirow{3}{*}{E2E} &
  ViLT &
  \textbf{80.36} &
  \textbf{86.45} &
  86.35 &
  \textbf{71.40} &
  70.26 &
  80.00 & 83.45 &  \textbf{81.90} &  \textbf{78.67} & 81.60 &  \textbf{83.20} &  \textbf{82.20} &
  \multirow{3}{*}{\textbf{79.32}} \\
     & ALBEF  & 79.33 & 84.20 & \textbf{87.30} & 69.45 & 72.08 & 81.65 &  \textbf{86.65} & 77.05 & 71.13 & 85.50 & 80.00 & 78.05        \\
 &
  TCL &
  78.27 &
  83.00 &
  85.20 &
  63.45 &
  \textbf{73.10} &
  \textbf{82.85} & 85.40 & 76.65 & 71.47 &  \textbf{83.50} & 79.75 & 77.40
   \\ \hline
DUAL & CLIP   & 67.93 & 72.90 & 73.60 & 65.20 & 63.74 & 74.25 & 69.30 & 65.90 & 61.33 & 69.85 & 67.20 & 62.00 & 67.93 \\ \hline
\end{tabular}
}
    \caption{Evaluation of Attribute}
    \label{attr}
\end{table*}

\section{Results and Analysis}


Table \ref{all} shows the overall performance of the compared models on Object, Relation, and Attribute.  The results indicate that understanding Relation and Attribute is more challenging than Object. The ViLT achieved the best score overall. We describe the experimental results and the detailed analysis of the seven representative VLP models based on the proposed framework.



\subsection{Evaluation of Object}

\begin{figure*}
\centering
\subfigure{
\begin{minipage}[t]{0.20\linewidth}
\centering
\includegraphics[width=1.0in]{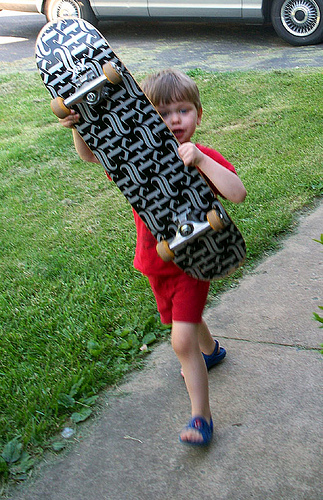}\\
\label{heat-map-car-a}
\end{minipage}%
}%
\subfigure{
\begin{minipage}[t]{0.20\linewidth}
\centering
\includegraphics[width=1.0in]{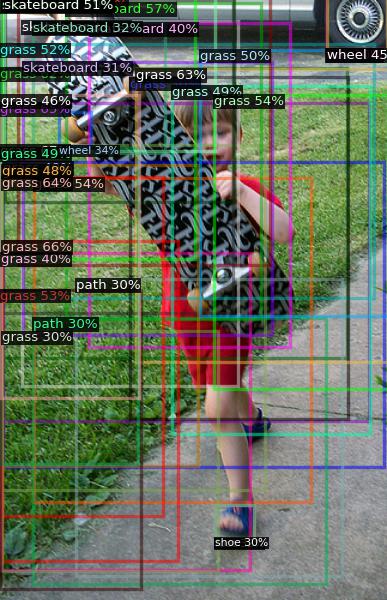}\\
\label{heat-map-car-b}
\end{minipage}%
}%
\subfigure{
\begin{minipage}[t]{0.20\linewidth}
\centering
\includegraphics[width=1.0in]{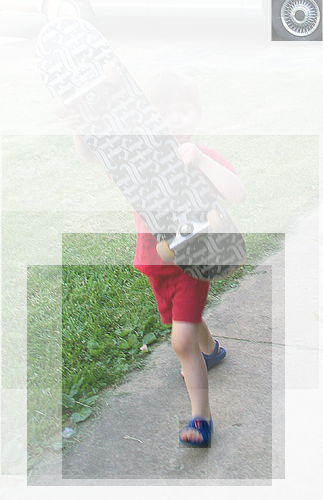}\\
\label{heat-map-car-c}
\end{minipage}
}%
\subfigure{
\begin{minipage}[t]{0.20\linewidth}
\centering
\includegraphics[width=1.0in]{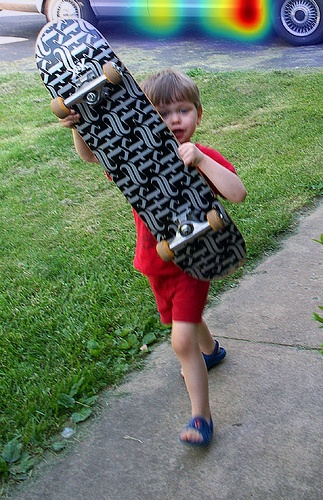}\\
\label{heat-map-car-d}
\end{minipage}
}%
\centering
\caption{\textbf{(a)} The original image, its descriptive text is: 'wheel ON \textit{car}'. \textbf{(b)} The bounding boxes extracted by Faster-RCNN, which fail to detect \textit{car} correctly. \textbf{(c)} The heat-map of LXMERT for the word \textit{car}, which misses the true position. \textbf{(d)} The heat-map of ALBEF for the word \textit{car}, which focus on the true position.}
\label{heat-map-car}

\end{figure*}

Object size and location greatly influence performance (Table~\ref{obj}). Specifically, we observe that all models gain the highest scores in Large and Center, which shows that VLP models tend to focus on large objects and those in the central point.
Particularly, CLIP has the most performance degradation when it’s moved from Large to Small and Center to Margin (13.12\% and 12.01\% degradation, respectively). Moreover, end-to-end (E2E) models (ViLT, ALBEF, and TCL) are more robust to Location variance than others. On average, their scores only drop 2.82\% from Center to Margin, compared with an average of 6.44\% for region-based models (OSCAR, UNITER, and LXMERT) and 12.01\% for CLIP.
The performance gap between the Center and Margin indicates that E2E models are more stable than region-based models. We speculate the reason to be the region-based models are limited to the performance of object detector(\textit{e.g.} Faster R-CNN), whose detection performance may drop for some areas of the images. Moreover, object detectors tend to ignore incomplete objects located at the margin of an image so that these marginal objects will not be fed into the cross-modal transformer. Therefore, the region-based models are prone to miss some useful information located in the margin. Conversely, the E2E model has the flexibility to focus on the whole image area (Figure \ref{heat-map-car}). The object detector cannot identify the \textit{car} on the top of the image since the bounding box does not contain information about the \textit{car}, as in Figure \ref{heat-map-car}(c). In contrast to the region-based model, the E2E model correctly focuses on the area of the \textit{car}, as in Figure \ref{heat-map-car}(d).


\begin{figure*}
\centering
\subfigure{
\begin{minipage}[t]{0.23\linewidth}
\centering
\includegraphics[width=1.2in]{images/heatmap/working_oscar.png}\\
\label{heat-map-working-a}
\end{minipage}%
}%
\subfigure{
\begin{minipage}[t]{0.23\linewidth}
\centering
\includegraphics[width=1.2in]{images/heatmap/working_lxmert.png}\\
\label{heat-map-working-b}
\end{minipage}%
}%
\subfigure{
\begin{minipage}[t]{0.23\linewidth}
\centering
\includegraphics[width=1.2in]{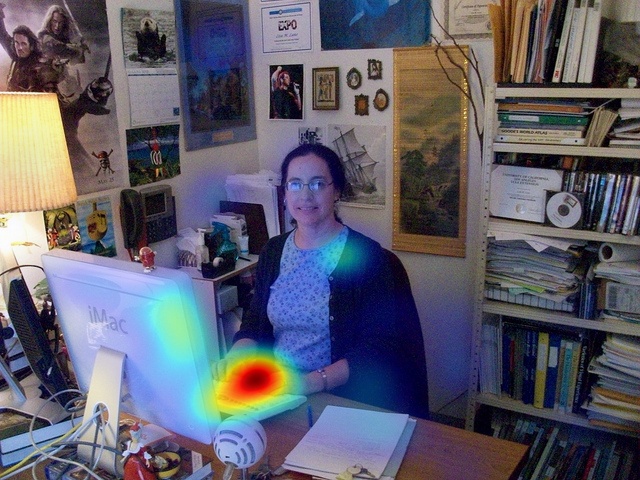}\\
\label{heat-map-working-c}
\end{minipage}
}%
\subfigure{
\begin{minipage}[t]{0.23\linewidth}
\centering
\includegraphics[width=1.2in]{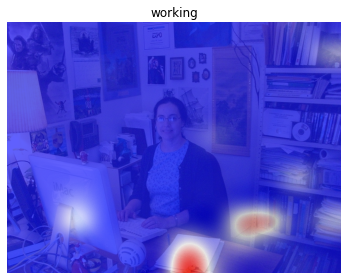}
\label{heat-map-working-d}
\end{minipage}
}%
\centering
\caption{The heat-map for the word '\textit{working}' of OSCAR, LXMERT, ALBEF and ViLT, respectively}
\label{fig:heat-map-working}
\end{figure*}
\subsection{Evaluation of Relation}

Aligning with the previous research \cite{hendricks2021probing}, the performance on Relation is generally low for all models (Table \ref{relation}), indicating that detecting relation is a more challenging problem than object and attribute recognition. 
We found that the CLIP performs the best (74.07\%) in understanding Action but achieves the worst on the Spatial (49.1\%). \citet{subramanian2022reclip} presented a similar observation that the CLIP was challenging to recognize Spatial expressions. 
The size and location of the objects are less critical to Relation changes.  
Region-based models drop 6.44\% from Center to Margin when evaluating on Object, while their performance degradation about Location is reduced to 3.44\% when evaluating on Relation. It indicates that the difficulty of relation recognition is not closely related to image level changes (\textit{i.e.} Location and Size) but is closely related to text level information. 


Additionally, we generate heat-maps to show where VLP models focus on Relation (Figure \ref{fig:heat-map-working}). The caption on the image is 'the woman is \textit{working} on her computer at the desk', and we generate the heat-maps of four different models for the word \textit{working}. The clear evidence of the word \textit{working} is where the woman is typing on the keyboard with her hand. OSCAR and ALBEF have reasonable attention maps, while LXMERT has less attention on the hands and keyboard. ViLT has the attention map on the paper and books, but not on the person.

\subsection{Evaluation of Attribute}  
We explore the performance of attribute recognition on the location and size of objects. The average score of the E2E model is 9.42\% higher than the region-based model on both image and text levels (Table \ref{attr}). The dual-architecture model is 1.97\% worse than the region-based model on average and has an 11.39\% degradation compared to the end-to-end model. All models get relatively high scores on Color and Material, low on Size, State, and Action, and perform the worst on Size (Table \ref{attr}). 
Understanding attribute requires complex semantic cognition between a text and an image, often beyond simple visual recognition~\cite{zellers2019recognition}. For example, Material is not only a matter of shape, size, and color but also something beyond optical recognition. When humans look at the image (e.g., artificial turf), it is hard to tell whether natural glass or artificial turf. 

The challenge in understanding Size is mainly due to the subjectivity in natural language. For example, the size in natural language does not translate to an absolute pixel size in images~\cite{zhang2022visual}. During pretraining, the VLP models are trained on various datasets where size words (e.g., small, large, giant) may not be consistent with each other in terms of visual Size. Also, the expressions of object sizes are different depending on the angles of a camera, the background, or an annotation's personal view. The performance of Action is also low because some actions are impossible to judge without any sequence of the frames (e.g., open door vs. close door). Therefore, research on the video-language multi-modal has been actively explored~\cite{wang2022all}. 




 
 

\section{Conclusion}
In conclusion, this paper introduces VL-CheckList to analyze VLP models from three perspectives: object, attribute, and relation. We adopt four corpora with carefully constructed taxonomy to produce fine-grained benchmark results for seven popular VLP models. Experiments not only validate the proposed method's soundness but also provide several valuable insights about these methods, e.g., the size and location of the objects are critical, and relation recognition is one of the most challenging in the VLP models. 
In the future, we plan to include more fine-grained evaluation aspects into VL-CheckList and improve existing VLP methods to achieve state-of-the-art performance under the guidance of the VL-CheckList report.

\section{Limitations}
The goal of the VL-CheckList is to quickly evaluate the robustness on the linguistic variations. Thus, we choose the ITM, which is universal and can easily test a model without finetuning each task. However, we have not yet comprehensively investigated why the high ITM-scored models sometimes achieve lower scores on the downstream tasks in this work. For a certain task, how it is finetuned is more critical to model performance. We leave a detailed study of them to future work.


\bibliography{anthology,custom}
\bibliographystyle{aaai23}


\end{document}